\documentclass{article}

\usepackage{arxiv}

\usepackage[utf8]{inputenc}
\usepackage[T1]{fontenc}    % use 8-bit T1 fonts
\usepackage{hyperref}       % hyperlinks
\usepackage{url}            % simple URL typesetting
\usepackage{booktabs}       % professional-quality tables
\usepackage{amsfonts}       % blackboard math symbols
\usepackage{nicefrac}       % compact symbols for 1/2, etc.
\usepackage{microtype}      % microtypography
\usepackage{comment}
\usepackage{lipsum}
\usepackage{graphicx}
\graphicspath{ {./images/} }
\usepackage[ruled,vlined]{algorithm2e}
\usepackage{amsmath,amssymb}
\DeclareMathOperator*{\argmax}{arg\,max}
\DeclareMathOperator*{\argmin}{arg\,min}

\title{Generalized Planning With Deep Reinforcement Learning}
\author{
 Or Rivlin \\
  Technion Autonomous Systems Program\\
  Technion - Israel Institute of Technology\\
  Haifa 3200003 \\
  \texttt{srivlin@campus.technion.ac.il} \\
  \And
 Tamir Hazan \\
  Faculty of Industrial Engineering and Management\\
  Technion - Israel Institute of Technology\\
  Haifa 3200003 \\
  \texttt{tamir.hazan@technion.ac.il} \\
    \And
 Erez Karpas \\
  Faculty of Industrial Engineering and Management\\
  Technion - Israel Institute of Technology\\
  Haifa 3200003 \\
  \texttt{karpase@technion.ac.il} \\
}
\date{December 2019}

\usepackage{graphicx}

\begin{document}

\maketitle

\begin{abstract}
    A hallmark of intelligence is the ability to deduce general principles from examples, which are correct beyond the range of those observed. Generalized Planning deals with finding such principles for a class of planning problems, so that principles discovered using small instances of a domain can be used to solve much larger instances of the same domain. In this work we study the use of Deep Reinforcement Learning and Graph Neural Networks to learn such generalized policies and demonstrate that they can generalize to instances that are orders of magnitude larger than those they were trained on.
\end{abstract}

\section{Introduction}

Classical Planning is concerned with finding plans, or sequences of actions, that when applied to some initial condition specified by a set of logical predicates, will bring the environment to a state that satisfies a set of goal predicates. This is usually performed by some heuristic search procedure, and the resulting plan is applicable only to the specific instance that was solved. However, a possibly stronger outcome would be to find some sort of higher level plan that can solve many instances that belong to the same domain, and thus share an underlying structure. The study of methods that can discover such higher level plans is called Generalized Planning. Generalized plans do not necessarily exist for all classical planning domains, but finding such solutions for domains in which it is possible could obviate the need to perform compute intensive search in cases where we only wish to find a goal satisfying solution.
To give an example of such a generalized plan, let us consider a simplified Blocksworld domain. In this domain there are unique blocks that can be either stacked on each other or strewn about the floor, and the goal is to stack and unstack blocks such that we arrive at a goal configuration from an initial configuration. Finding a plan that does so in an optimal number of steps is generally NP-hard \cite{gupta1992complexity}, but finding a plan that satisfies the goal regardless of cost can be done in polynomial time in the following manner:
\begin{enumerate}
  \item Unstack all the blocks so that they are scattered on the floor
  \item stack the block according to the goal configuration, beginning with the lower blocks
\end{enumerate}

This strategy is not optimal since we might unstack blocks that are already in their proper place according to the goal specification, but it will yield a goal satisfying plan for every instance in this simplified Blocksworld domain. Such a generalized strategy can also be thought of as a policy, which raises the possibility of learning it through reinforcement learning. Machine learning theory often assumes that our training data distribution is representative of the test data distribution, thus justifying our expectation that our models generalize well to the test data. In generalized planning this is not the case, as our test instances could be much larger than the training instances, and thus far out of the training distribution. In this work we show that having the \textbf{right} inductive bias in the form of a neural network architecture could lead to models that effectively learn policies that are akin to general principles, and can solve problems that are orders of magnitude larger than those encountered during training.

\section{Background}

\subsection{Classical Planning}

Classical planning uses a formal description language called Planning Domain Definition Language (PDDL) \cite{mcdermott1998pddl}, derived from the STRIPS modeling language \cite{fikes1971strips} to define problem domains and their corresponding states and goals. we are concerned with satisficing planning tasks, which can be defined by a set $(F,O,I,G)$ where $F$ is a set of propositions (or predicates) that describe the properties of the objects present in task instance and their relations, $O$ is a set of operators (or actions types), $I \subseteq F$ is the initial state and $G \subseteq F$ is a set of goal states. each action type $o \in O$ is defined by a triple $(Pre(o), Add(o), Del(o))$, where the preconditions $Pre(o)$ is a set of predicates that must have a true value for the action to be applicable, $Add(o)$  is a set of predicates which the action turns to true upon application and $Del(o)$ is a set of predicates which the action turns false upon application. We seek to find a plan, or a sequence of actions that when applied will lead to a state $s$ for which $G \subseteq s$, within some time limit or a predefined number of steps. Finding plans for planning tasks is often accomplished by heuristic search methods, however in this work we focus on learning reactive planning policies that can train on instances of a specific domain and then generalize to new, unseen instances in that same domain.

\subsection{Reinforcement Learning}

Reinforcement learning (RL) is a branch of machine learning that deals with learning policies for sequential decision making problems. RL algorithms most often assume the problem can be modelled as a Markov Decision Process (MDP), which in the finite horizon case is defined by a tuple ($S$, $A$, $R$, $P$, $T$, $\rho$), where $S$ is the set of states, $A$ is the set of actions, $R$ is a reward function that maps states or state-actions to some scalar reward, $P$ is the transition probability function such that $p(s'|s,a)=P(s',s,a)$, $T$ is the task horizon and $\rho$ is the distribution over initial states. The value of a policy $\pi$ in the finite horizon RL problem is:
\begin{equation}
J(\pi) = \displaystyle \mathbb{E}_{\tau \sim \pi} \left[ \sum_{t=0}^{T} r(s_{t},a_{t}) \right]
\end{equation}
Where $\tau$ are trajectories sampled by the distribution induced by the policy $\pi$, initial state distribution $\rho$ and transition function $P$, and $r(s_{t},a_{t})$ is the reward received after taking action $a_{t}$ at state $s_{t}$. The learning problem can thus be formalized as an optimization problem, in which we wish to find the best policy:
\begin{equation}
\pi ^* =  \argmax_{\pi} J(\pi)
\end{equation}

In the case of large state and action spaces, we cannot hope to represent our policy as a table, and are thus forced to use function approximators to represent the policy, with some parameters $\theta$. We focus on stochastic policies, which map states and actions to probabilities, such that $p(a|s) = \pi_{\theta}(a|s)$, and use policy gradient based methods to optimize our policies \cite{williams1992simple}. Policy gradient methods estimate the gradient of the objective function with respect to the policy parameters using monte-carlo sampling. The gradient of the RL objective is:
\begin{equation}
\nabla_{\theta}J(\pi_{\theta}) = \displaystyle \mathbb{E}_{\tau \sim \pi} \left[ \sum_{t=0}^{T} \nabla_{\pi} \log \pi_{\theta}(a_{t}|s_{t}) \sum_{t^{'}=t}^{T} r(s_{t^{'}},a_{t^{'}}) \right] = \displaystyle \mathbb{E}_{\tau \sim \pi} \left[ \sum_{t=0}^{T} \nabla_{\pi} \log \pi_{\theta}(a_{t}|s_{t}) R(t) \right]
\end{equation}

Where $R(t)$ is the "return to-go". When implementing policy gradient methods, we can estimate the policy gradient by taking the gradient of a "pseudo-loss", computed using sampled trajectories:
\begin{equation}
L_{pseudo} = \frac{1}{|D_{k}|} \sum_{\tau \in D_{k}} \sum_{t=0}^{T} \log \pi_{\theta}(a_{t}|s_{t}) R(t)
\end{equation}
Where $D_{k}$ is a collection of trajectories sampled at iteration $k$ of the algorithm. We can optimize our policy by gradient ascent using the following equation:
\begin{equation}
\pi_{\theta_{k+1}} = \pi_{\theta_{k}} + \alpha \nabla_{\theta}L_{pseudo}
\end{equation}

This kind of algorithm is "on-policy", which means that data used to update the policy must be generated by the same policy parameters. This requires the algorithm to discard all the data it gathered after each update and collect new data for the next update, which makes on-policy algorithms data inefficient.

\subsection{Proximal Policy Optimization}

Proximal Policy Optimization (PPO) \cite{schulman2017proximal} is a policy gradient based algorithm that seeks to better exploit the data gathered during the learning process, by performing several gradient updates on the collected data before discarding it to collect more. In order to avoid stability issues that could arise from large policy updates, PPO uses a special clipped objective to discourage divergence between the current policy  and the data collection policy, to define the following optimization problem:

\begin{equation}
\theta_{k+1} = \argmax_{\theta} \frac{1}{|D_{k}|} \sum_{\tau \in D_{k}} \sum_{t=0}^{T} max \left( \frac{\pi_{\theta}(a_{t}|s_{t})}{\pi_{\theta_{k}}(a_{t}|s_{t})} A(a_{t}|s_{t})^{\pi_{\theta_{k}}}, clip \left(\frac{\pi_{\theta}(a_{t}|s_{t})}{\pi_{\theta_{k}}(a_{t}|s_{t})}, 1-\epsilon, 1+\epsilon \right) A(a_{t}|s_{t})^{\pi_{\theta_{k}}} \right) 
\end{equation}

Where $clip$ is a function that clips the values of its input to be between the specified minimum and maximum values, $\pi_{\theta}$ is the policy we are currently optimizing, $\pi_{\theta_{k}}$ is the policy used to collect the data (before updating) and $A(a_{t}|s_{t})^{\pi_{\theta_{k}}}$ is the advantage of the action given the current state and parameters:
\begin{equation}
A(a_{t}|s_{t})^{\pi_{\theta_{k}}} = R(s_{t}|\theta_{k}) - V_{\phi_{k}}(s)
\end{equation}

Where the dependency on the action $a$ comes from the empirical return to-go $R(s_{t})$, which depends on the specific actions that were taken by the policy. $V_{\phi_{k}}(s)$ is a state value predicted by some function approximator with parameters $\phi_{k}$, obtained at each iteration by solving:

\begin{equation}
\phi_{k+1} = \argmin_{\phi} \frac{1}{|D_{k}|} \sum_{\tau \in D_{k}} \sum_{t=0}^{T}  \left| V_{\phi}(s_{t}) - R(s_{t}) \right| ^{2} 
\end{equation}

\section{Learning Generalized Policies}
\subsection{State Representation}
We chose to represent the states in our framework as graphs, with features encoding the properties and relations between the objects in a given state. Our framework operates on problem domains specified by the PDDL modeling language, in which problem instances are defined by a list of objects and a list of predicates that describe the properties of these objects and the relations between them at the current state. We limit ourselves to domains for which predicates have an arity of no more than two, which is not a significant limitation since higher arity predicates can in many cases be decomposed to several lower arity predicates. 
Our graphs are composed of global features, node features and edge features, as in \cite{battaglia2018relational}. We denote our global features $U$, our nodes $V$ and our edges $E$. Global features represent properties of the problem instance or entities that are unique for the domain, such as the hand in the Blocksworld domain, and are determined by the 0-arity predicates of the domain. Node features represent properties of the objects in the domain, such as their type, and are determined by the 1-arity predicates. Lastly, edge features represent relations between the objects and are determined by the 2-arity predicates. \\
When producing a graph representation of a PDDL instance state, a complete graph is produced with a node for each object in the state. For each predicate in the state, the corresponding feature is assigned a binary value of 1, and all other features are assumed to be false with a value of 0. In order to include the goal configuration in the input to the neural network, the goal predicates are treated almost as if they were another state-graph, and the two graphs are concatenated together to form a single representation for the state-goal. The difference between state graphs and goal graphs, is that in the goal graphs a 0 valued feature means that it contributes no goal, and in the state graph a 0 valued feature means that the predicate is assigned a false value.
\par

\begin{figure}[h!]
    \begin{center}
        \includegraphics[width=0.9\textwidth]{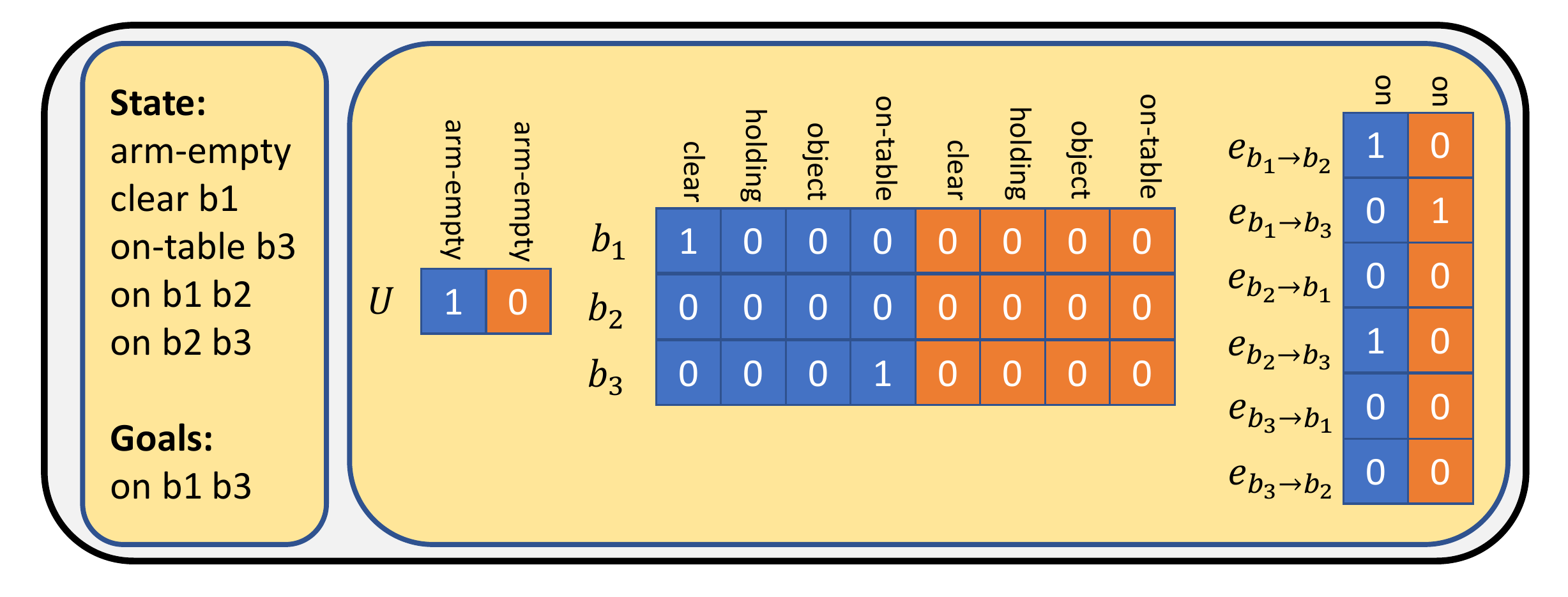}
        \caption{Example state-goal graph from the Blocksworld domain. The left side shows the PDDL description of the state-goal, and the right side shows the graph representation of the same state-goal. Blue represents state information and orange represents goal information}
        \label{fig:graphexample}
    \end{center}
\end{figure}

\begin{figure}[h!]
    \begin{center}
        \includegraphics[width=0.6\textwidth]{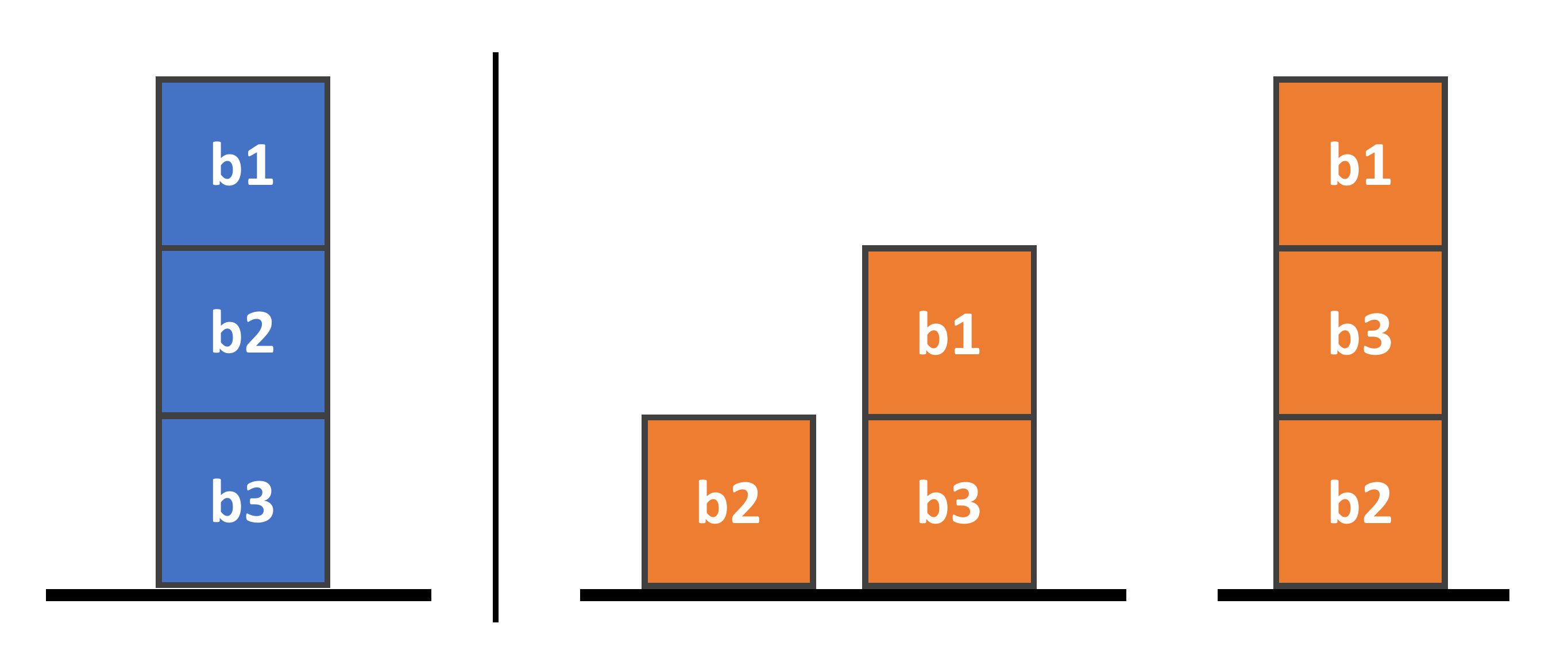}
        \caption{Visualization of the state from the previous figure (blue) and possible goal-satisfying configurations (orange)}
        \label{fig:stategoalsexample}
    \end{center}
\end{figure}

The classical planning domains used throughout this work are deterministic and Markovian, meaning that the current state holds all the required information to solve the problem optimally. Despite this property, we found that adding past states in addition to the current one helps the learning process and improves the generalization capability to larger instances. While this is not strictly essential, our experiments suggest that this step helps the policy mitigate "back-and-forth" behavior to some extent, and this is especially helpful on the larger instances where the policy is more prone to make mistakes and then attempt to correct them. Adding this history is straight-forward; we simply concatenate the graphs for the K previous states and current state, and then concatenate the goal graph as mentioned previously. We tested several such history horizons, and found that adding only the last state results in overall best performance and generalization. An example of a state-goal graph from the Blocksworld domain can be seen in figures \ref{fig:graphexample} and \ref{fig:stategoalsexample}, showing an instance with 3 blocks.

\subsection{Graph Embedding}

In order to learn good policies using the graph representations of state-goals we first use a Graph Neural Network (GNN) to embed the node, edge and global features of the graph in respective latent spaces. The GNN performs message passing between the different components of the graph, allowing useful information to flow. We use two different types of GNN blocks, each enforces a different style of information flow within the graph and thus more suited to certain problem domains than others. In both of these types the update order is similar and takes the following common form:
\begin{enumerate}
  \item Edges are updated using the previous edges and the "origin" nodes of those edges.
  \item Nodes are updated using the previous nodes, the incoming updated edges and the global features.
  \item Globals are updated using the previous globals and the aggregation of the updated nodes.
\end{enumerate}

\begin{figure}[h!]
    \begin{center}
        \includegraphics[width=0.9\textwidth]{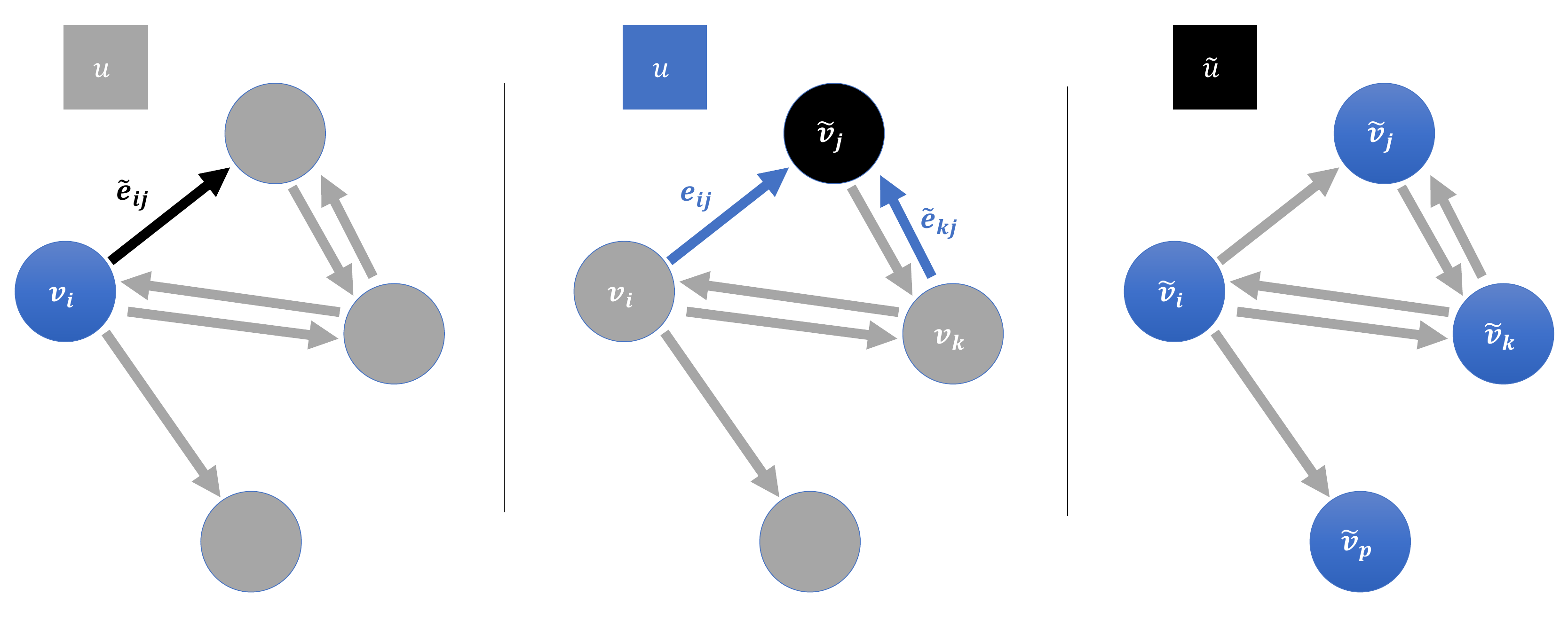}
        \caption{Flow of information within the graph network block. The black piece is the component being updated at each step, the additional information used to update that component is in blue color, and the gray information is not used. On the left: updating the edges. In the middle: updating the nodes. On the right: updating the global features}
        \label{fig:graphflow}
    \end{center}
\end{figure}

The first block type we used is similar to the one described in \cite{battaglia2018relational} which we name accordingly \textbf{Graph Network block} (GN block). Mathematically, this block performs the following operations:

\begin{equation}
\tilde{e}_{ij} = \phi(W^{e}[e_{ij}, v_i] + b^{e})
\end{equation}
\begin{equation}
h_{ij} = \phi(W^{v}_1[V_i, \tilde{e}_{ij}] + b^{v}_1)
\end{equation}
\begin{equation}
m_{i} = \psi(h_{ij})
\end{equation}
\begin{equation}
\tilde{v}_{i} = \phi(W^{v}_2[h_{i}, u] + b^{v}_2)
\end{equation}
\begin{equation}
\tilde{u} = \phi(W^{u}\frac{1}{\arrowvert v \arrowvert}\sum_{i\in{v}}\tilde{v}_{i} + b^{u})
\end{equation}

In the above notation, $\phi$ is a nonlinearity such as Rectified Linear Unit, $\psi$ is a node-wise max-pooling operation and $W$, $b$ are respective weight matrices and biases. In the GN block, nodes receive messages from their neighbouring nodes indiscriminately, which works well to propagate general information across the graph but makes it harder to transfer specific bits of information when needed.
\par
The second type of block was designed to address that shortcoming of the GN block, and for that purpose was endowed with an attention mechanism. We named the second block \textbf{Graph Network Attention block} (GNAT block), and unlike the Graph Attention Network of \cite{velivckovic2017graph}, it uses an attention mechanism similar to the Transformer model of \cite{vaswani2017attention}. This block performs the following operations:

\begin{equation}
\tilde{e}_{ij} = \phi(W^{e}[e_{ij}, v_i] + b^{e})
\end{equation}
\begin{equation}
h_{i} = \phi(W^{v}_1 v_{i} + b^{v}_1)
\end{equation}
\begin{equation}
k_{i} = \phi(W^{k} v_{i} + b^{k})
\end{equation}
\begin{equation}
q_{ij} = \phi(W^{q} \tilde{e}_{ij} + b^{q})
\end{equation}
\begin{equation}
%\alpha_{ij} = Softmax(\sum_{columns}(K_{i}\odot Q_{ij}))
%\alpha_{ij} = Softmax(K_{i}^{T}Q_{ij})
\alpha_{ij} = \frac{\exp{(k_{i}^{T}q_{ij}})}{\sum_{p} \exp{(k_{i}^{T}q_{ip}})}
\end{equation}
\begin{equation}
m_{i} = \varphi(\alpha_{ij}\cdot \tilde{e}_{ij})
\end{equation}
\begin{equation}
\tilde{v}_{i} = \phi(W^{v}[v_{i}, m_{i}, u] + b^{v})
\end{equation}
\begin{equation}
\tilde{u} = \phi(W^{u}\frac{1}{\arrowvert v \arrowvert}\sum_{i\in{v}}\tilde{v}_{i} + b^{u})
\end{equation}

In the above notation, $\varphi$ is a node-wise summation operation, $\odot$ is the Hadamard product and $W$, $b$ are respective weight matrices and biases. As mentioned above, this type of block allows certain bits of information to travel in the graph in a more deliberate manner, by endowing the nodes with the ability to focus on specific messages. When constructing our GNN model, we can stack several blocks of these types (and combinations of them) to attain a deeper graph embedding capacity. In most of our experiments we used two blocks, either two successive GN blocks, or  a GNAT block followed by a GN block. Each configuration excelled at a different group of problems as we will show in the experiments section.

\subsection{Policy Representation}

Unlike common reinforcement learning benchmarks where the set of actions is fixed and can be conveniently handled by standard neural network architectures, in classical planning problems the set of actions is state dependent and varies in size between states. In PDDL, each domain description defines a set of action types that can be instantiated by grounding those action types to the state. Each action type receives a set of arguments, and in order to be applicable the arguments of the action must conform to a set of pre-conditions. For example, the Blocksworld domain has an action type called "pick-up" which gets a single block object as an argument. This block must be "clear", "on-table" and the "arm-empty" property must be true for the action to be applicable. All blocks that comply with these pre-conditions can be picked up, and represent a unique action. In addition to pre-conditions, each action type also has effects which are caused to the states upon application of the action. Some of these effects could be positive (certain predicates of the state will take a true value) and some negative (predicates will assume a false value).
\par
\begin{figure}[h!]
    \begin{center}
        \includegraphics[width=0.9\textwidth]{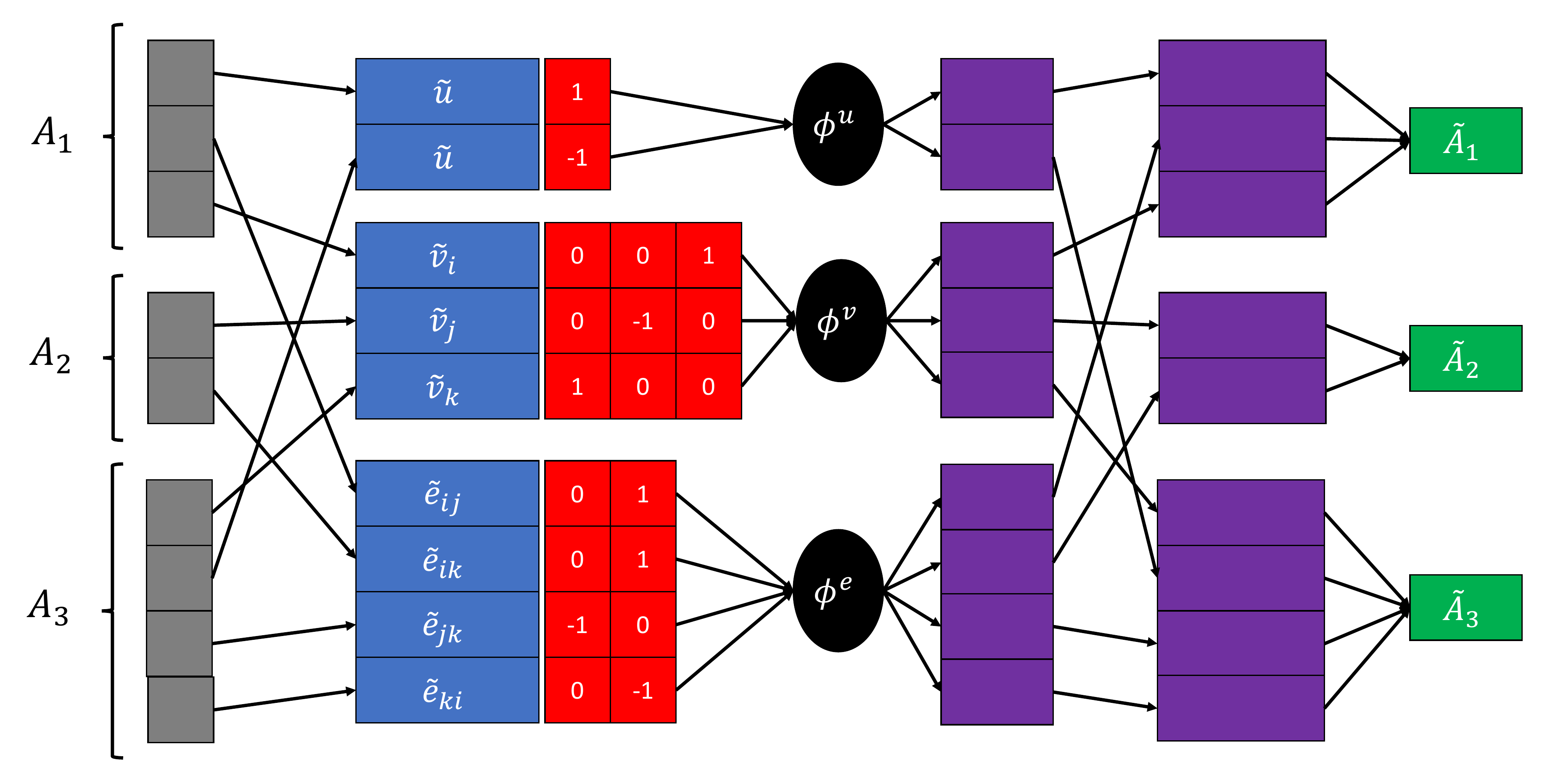}
        %\caption{Illustration of the action embedding process. The effects of the various actions are clustered to global effects, node effects %and edge effects, transformed and then placed back in their respective actions. At last, the effects are aggregated per-action to form a %single vector representation of the action}
        \caption{Illustration of the action embedding process. From left to right: gray blocks represent original action effects, blue blocks represent embedding of appropriate graph elements clustered (according to globals, nodes and edges), red blocks represent the one-hot effect type vectors, black ellipses represent MLPs, purple blocks represent effect embeddings clustered and then scattered back to their origin actions, and green blocks represent final action embeddings}
        \label{fig:actionschema}
    \end{center}
\end{figure}

At each step of planning, the successor-state generator gives the current state and a list of applicable actions. In order to represent the actions in a meaningful way that enables learning a policy over them, we chose to describe the actions in terms of their effects, since these are the essential components needed to make decisions. Since the successor-state generator provides the agent with all the legal actions at each step, we ignored the preconditions (all legal actions satisfy the pre-conditions). Each action is composed of several effects, each concerning a different aspect of the state, and are either positive or negative. The effects are clustered together based on their type (global effect, node effect or edge effect), and are represented as a concatenation of the embedding of the respective component and a one-hot vector describing which predicate is changed and if it is positive or negative. This one-hot vector is in the dimension of corresponding input component ($d_{v}$ for node effects for example) and contains either 1 for positive effects or -1 for negative effects at the appropriate predicate location. Each effect is transformed by a multi layered perceptron (MLP) according the its type and then the transformed effects are scattered back to their origin actions. The effects of each action are aggregated together to form a single vector representation of that action, which is fed eventually to the policy neural network. Figure \ref{fig:actionschema} illustrates the process of action representation.

The final policy is a MLP that outputs a single scalar for each action, and these scalars are then normalized by a softmax operation to get a discrete distribution over the actions. In addition, another MLP takes the final global feature embeddings of the graph and outputs the predicted value of the state, to be used for advantage estimation in the RL algorithm.

\subsection{Training Procedure}

Since the focus of this work was finding feasible plans, we chose to model our problem as a sparse reward problem with a binary reward. If the agent satisfies all the goals within a predefined horizon length, it gets a reward of 1, and if not it gets no reward. To determine an appropriate time limit we used the commonly used \textbf{hff} heuristic \cite{hoffmann2001ff}, which solves a relaxed version of the problem in linear time (the relaxed problem has no negative effects). We take the length of the relaxed plan and multiply it by a constant factor of 5 to get the horizon length.
\par
To train our policy we chose to use Proximal Policy Optimization (PPO) \cite{schulman2017proximal} for its simplicity and good performance. To handle the problem of sparse rewards we initially experimented with using Hindsight Experience Replay DQN \cite{andrychowicz2017hindsight} due to its demonstrated ability to tackle sparse goal reaching problems, but found that it introduced a lot of bias and resulted in unsatisfactory performance. To allow our policy to learn from a sparse binary reward, we resorted to a simpler method; we generated each training episode from a distribution over instance sizes, which includes sizes small enough to be occasionally solved by a randomly initialized policy. Doing this allows the policy to progress to eventually solve all the instance sizes in the distribution, without the need for a manual tuning of a curriculum. Although setting this distribution needs to be done manually, we found it very easy and quick to do by simple trial and error with a random untrained neural network.
\par
We made several small adjustments to the standard PPO algorithm which improved performance in our case. Many RL algorithms implementations roll out the policy for a fixed number of steps before updating the model parameters, often terminating episodes before completion in the process and using methods such as Generalized Advantage Estimation \cite{schulman2015high} and bootstrapping value estimations to estimate the returns as in \cite{mnih2016asynchronous}. We found these elements to add unwanted bias to our learning process, and instead rolled out each episode until termination, using empirical returns instead of bootstrapped value estimates to compute advantages. We also found that using many roll-outs and large batch sizes helped stabilize the learning process and resulted in better final performance, and so we performed 100 episode roll-outs and used the resulting data to update the model parameters at each iteration of the learning algorithm.

\subsection{Planning During Inference}

To improve the ability of our generalized policies to use additional time during test, we use them within a search algorithm, as was done in many other works such as \cite{silver2016mastering}, \cite{anthony2017thinking}. This type of synthesis gained great success in zero sum games such as Go and Chess \cite{silver2017mastering}, where a deep neural network policy was used in conjunction with a Monte Carlo Tree Search algorithm, which prompted other authors to do the same even for non-game problems \cite{abe2019solving}. We take a different approach and design our search algorithm specifically for the case of deterministic planning problems with a strong reactive policy.
Our algorithm is based on the classic Greedy Best First Search (GBFS) algorithm, but augments it in several key ways. In standard GBFS, a search tree is constructed from the root node, and at each iteration, the node with the best heuristic estimate is extracted from the open list, expanded and its child nodes added to the open list, and this procedure is repeated until a goal node is found or until time is out. Our algorithm, which we name GBFS-GNN, performs a similar procedure, but uses the policy and value functions to compute a heuristic value for each node, and performs a full roll-out for each expanded node. The offspring of the expanded node are added to the open list, but the rest of the nodes encountered during the roll-out are not, to avoid rapid memory consumption growth in large problems.
Each node in our search tree represents a state-action pair, and we use the following heuristic estimate for each node:

\begin{equation}
%g(s,a) = \pi (a|s) \cdot V(s) \frac{1}{1 + H(\pi(\cdot | s))}
g(s,a) = \frac{\pi (a|s) \cdot V(s)}{1 + H(\pi(\cdot | s))}
\end{equation}

\begin{figure}[h!]
    \begin{center}
        \includegraphics[width=0.6\textwidth]{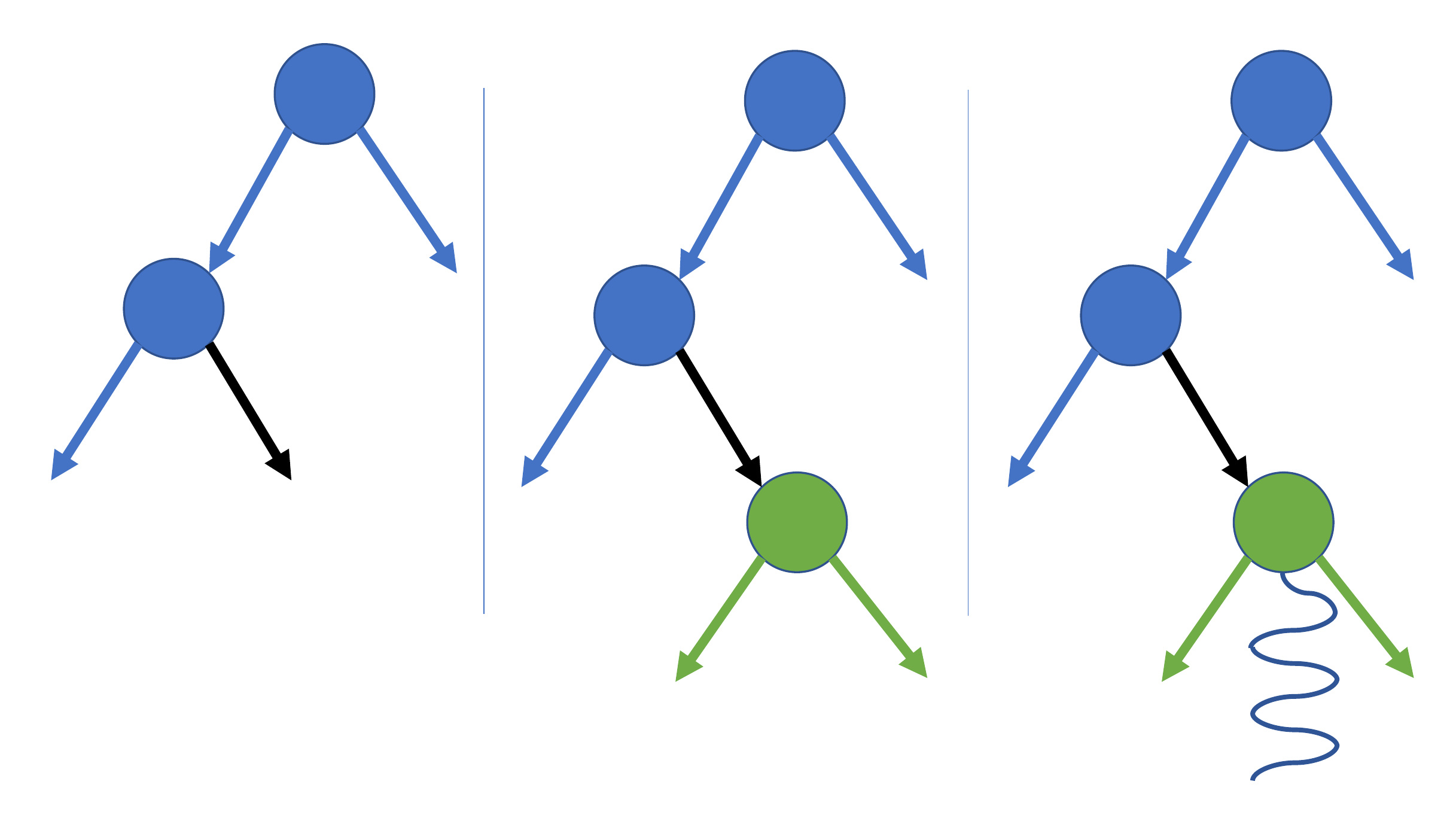}
        \caption{Illustration of the search procedure. From left to right: a state-action edge with the highest heuristic estimate is chosen (black), the action is applied to generate the successor state with its child edges (green) and finally a full roll-out is performed from the successor state using the policy (purple spiral)}
        \label{fig:gnngbfs}
    \end{center}
\end{figure}

In this equation, $g(s,a)$ is the heuristic estimate of the state-action, $\pi(a|s)$ is the probability of action $a$ under our policy $\pi$, $V(s)$ is the estimated state value according the the critic part of our neural network policy, and $H(\pi(\cdot | s))$ is the entropy of the policy's distribution over actions at state $s$. Figure \ref{fig:gnngbfs} illustrates our search algorithm.

\section{Related Work}

Learning to plan has been an active topic of research for many years, with different methods attempting to learn different aspects of a complete solver. Some works attempted to learn heuristic values of states for specific domains using features generated by other domain independent heuristics, such as \cite{yoon2006learning}, which learns heuristic values by regression. more recent works such as \cite{garrett2016learning} learn to rank successor states by using RankSVM \cite{joachims2002optimizing}. These types of methods do not explicitly use the state or goal information from the problem description, but rather learn using hand crafted features, and in addition do not learn an explicit planning policy over the available actions. Contrary to this, our methods learns planning policies over explicit states and goals, that directly choose actions to take.

Other works such as \cite{tamar2016value}, \cite{groshev2018learning} and \cite{guez2019investigation} learn an explicit planning policy over actions, using the actual state of the problem as input and a deep convolutional neural network, but rely on having a visual representation of the problem. This limits their usage to domains where a visual representation is available. Another limitation is that \cite{tamar2016value} and \cite{groshev2018learning} rely in addition on successful plans generated by a planning algorithm and learn policies using imitation learning, while \cite{guez2019investigation} use reinforcement learning for this purpose. Our work does not rely on visual representations or successful plans generated by planning algorithms, but learns directly from a PDDL representation of the problem by trial and error via deep reinforcement learning.

Some works have begun to study the use of graph representations of states and the use of different kinds of graph neural networks for the task of learning policies or heuristics. In \cite{toyer2018action} the authors proposed a unique kind of neural network called Action Schema Network (ASNet) which consists of alternating action layers and proposition layers to learn planning policies. They represent their state as a graph in which objects and actions are connected and propagate information back and forth to finally output a probability over actions. They train their ASNets by imitating plans generated by other planners, and augment the input with domain independent heuristic values to improve performance. In their experiments, they focus mainly on stochastic planning problems and demonstrate that their trained policies can generalize to larger instances than trained on. A limitation of ASNet is their fixed receptive field, that limits their ability to reason over long chains, which our work does not share.

In a recent paper, \cite{shen2019learning} propose an extension of \cite{battaglia2018relational} to hypergraphs, and use it to learn heuristics over hypergraphs that represent the delete relaxation states of planning problems. They use supervised learning of optimal heuristic values generated by a planning algorithm, and then use the resulting neural network as a heuristic function within a search algorithm. In contrast to this, our method focuses on learning policies, which can be more time-efficient during evaluation since a single forward pass over the neural network is needed to make a decision at each state. Using heuristic estimates requires estimating all the successor states of a state in order to choose the best action, which could potentially increase run-time. Another difference is that our work operates directly on states instead of delete relaxations, which might limit the power of heuristics since some information is omitted. Overviews of older methods of learning to plan can be found in \cite{minton2014machine}, \cite{zimmerman2003learning} and \cite{fern2011first}.

\section{Experiments}

\subsection{Domains}

We evaluate our approach on five common classical planning domains, Chosen from the IPC planning competition collection of domain generators that have predicates of arity no larger than 2: 
\begin{itemize}
  \item Blocksworld (4 op): A robotic arm must move blocks from an initial configuration in order to arrange them according to a goal configuration.
  \item Satellite: A fleet of satellites must take images of locations, each with a specified type of sensor.
  \item Logistics: Packages must be delivered to target locations, using airplanes and trucks to move them between cities and locations.
  \item Gripper: A twin-armed robot must deliver balls from room A to room B.
  \item Ferry: A ferry must transport cars from initial locations to designated target locations.
\end{itemize}
What these five domains have in common is that simple generalized plans can be formulated for them, which are capable of solving arbitrarily large instances. We wish to demonstrate that our method is capable of producing policies that solve much larger instances than those they were trained on, thus automatically discovering such generalized plans. Some domains are easier than others, and in cases where the generalized plan is very easy to describe we often witnessed that the policy generalizes very successfully. For example, the Gripper domain has a very simple strategy (Grab 2 balls with each trip to room B) and indeed our neural network learns the optimal strategy and usually still performs optimally even for instances with hundreds of balls.
To demonstrate that our policies indeed generalize well, we trained them on small instances and used both small and large instances for evaluation.
\begin{itemize}
\item For the Blocksworld domain we trained our policy on instances with 4 blocks, and evaluated on instances with 5-100 blocks. 
\item For the Satellite domain we trained our policy on instances with 1-3 satellites, 1-3 instruments per satellite, 1-3 types of instruments, 2-3 targets, and evaluated on instances with 1-14 satellites, 2-11 instruments per satellite, 1-6 types of instruments and 2-42 targets.
\item For the Logistics domain we trained our policy on instances with 2-3 airplanes, 2-3 cities, 2-3 locations per city, 1-2 packages, and evaluated on instances with 4-12 airplanes, 4-15 cities, 1-6 locations per city and 8-40 packages.
\item For the Gripper domain we trained our policy on instances with 3 balls, and evaluated on instances with 5-200 balls.
\item For the Ferry domain we trained our policy on instances with 3-4 locations, 2-3 cars and evaluated on instances with 4-40 locations and 2-120 cars.
\end{itemize}

\subsection{Experimental setting}

For training our policies, we rely on having instance generators to produce random training instances, since our method requires large amounts of training data. All policies are trained for 1000 iterations, each with 100 training episodes and up to 20 gradient update steps. Experiments are performed on a single machine with a i7-8700K processor and a single NVIDIA GTX 1070 GPU. We used the same training hyperparameters for all five domains, but slightly varying neural network models. We used a hidden representation size of 256 and ReLU activations, a learning rate of 0.0001, a discount factor of 0.99, an entropy bonus of 0.01, a clipping ratio of 0.2 and a KL divergence cutoff parameter of 0.01. For the Blocksworld and Gripper domains we used a two layer GNN with both layers of the GN block type, and for the Satellite, Ferry and Logistics domains we used a two layer GNN with a GNAT block followed by a GN block. Our code was implemented in Python and our neural networks and learning algorithm were implemented using PyTorch \cite{paszke2019pytorch}.

\subsection{Baselines}

We focus in our evaluation on solving large instances of generalized planning domains and compare our method with a classical planner. Other learning based methods either had no available code by the time this work was written (such as \cite{shen2019learning}) or were inherently limited in scaling to the large problems (for example \cite{toyer2018action}), so we opted for a more general baseline in the form of a classical planner, which can scale to large problems given enough time and memory. We compare against fast-downward \cite{helmert2006fast}, which is a state of the art framework. Our approach uses Pyperplan as the model and successor state generator, which is a Python based framework. We use the LAMA-first configuration as the setup for fast-downward, as it is a top performing competitive satisficing planning algorithm.

\subsection{Evaluation Metrics}

Since our work is focused on satisficing planning, we use success rate as our main metric. We run both our GBFS-GNN and fast-downward on a set of 50 held out evaluation instances per domain, and run each method for a fixed time limit of 600 seconds per instance, we then plot the success rate of each method against the time limit and against the number of expanded states to see how each method scales with given computation. The evaluation instances are generated according to a wide distribution such that both small and large instances are sampled.

\subsection{Results}

\begin{figure}[h!]
    \begin{center}
        \includegraphics[width=1.0\textwidth]{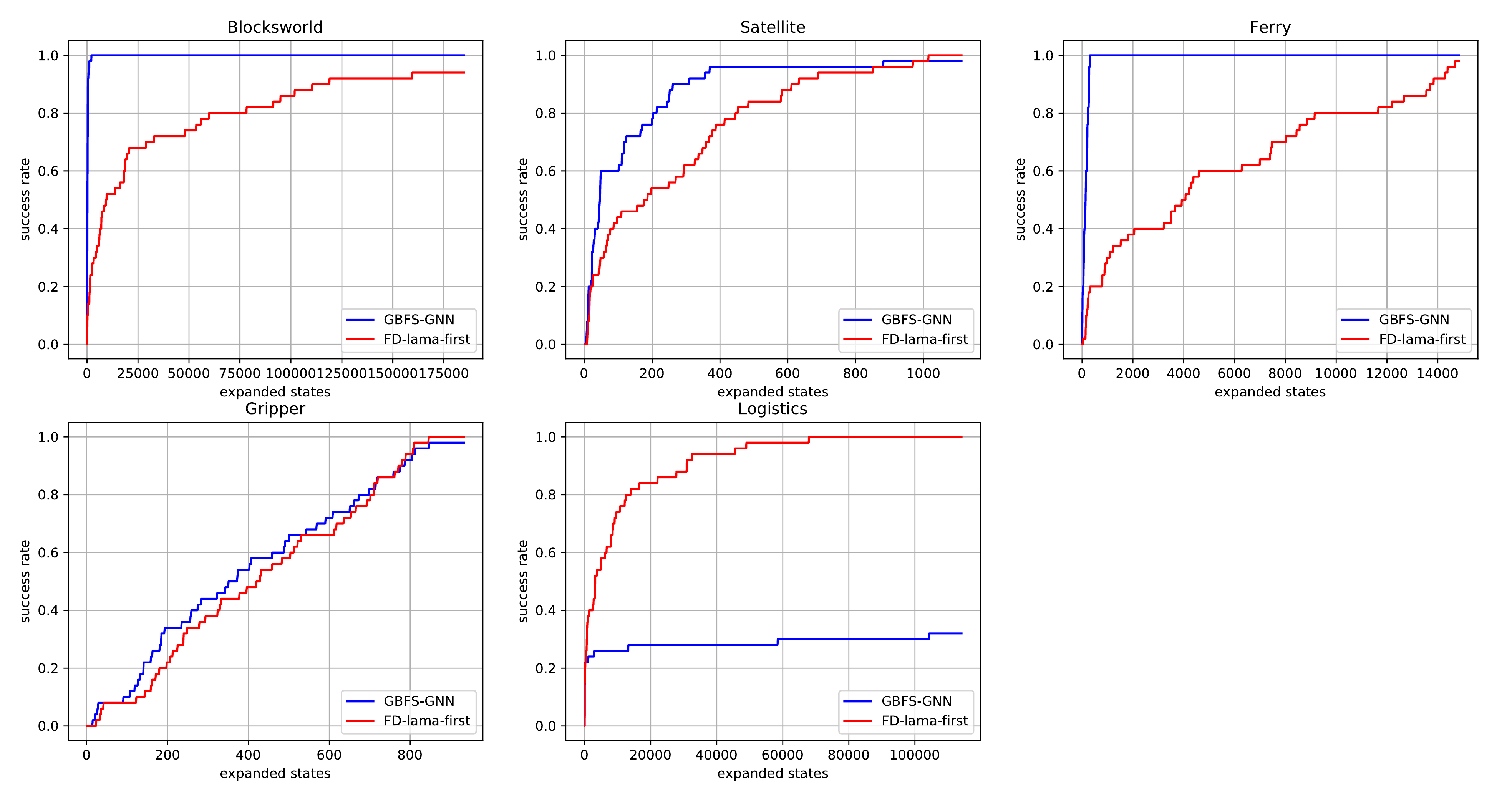}
        \caption{These plots compare success rate against number of expanded states for the various domains used in the evaluation}
        \label{fig:sr_exp_results}
    \end{center}
\end{figure}

We now present our results. Figure \ref{fig:sr_exp_results} shows a comparison between our method and fast-downward for the five domains we used in our experiments. The plots show success rate as a function of number of expanded states, and demonstrate that our method indeed scales favourably compared to the classical planner on 4 of the 5 domains. In fact, on the 4 domains where our policies generalized well, GBFS-GNN required very little to no search. In these domains, a solution can be found by just greedily following the policy in all but the hardest instances. Our search algorithm builds on this generalization capability and uses a small number of full policy roll-outs while searching.

\begin{figure}[h!]
    \begin{center}
        \includegraphics[width=1.0\textwidth]{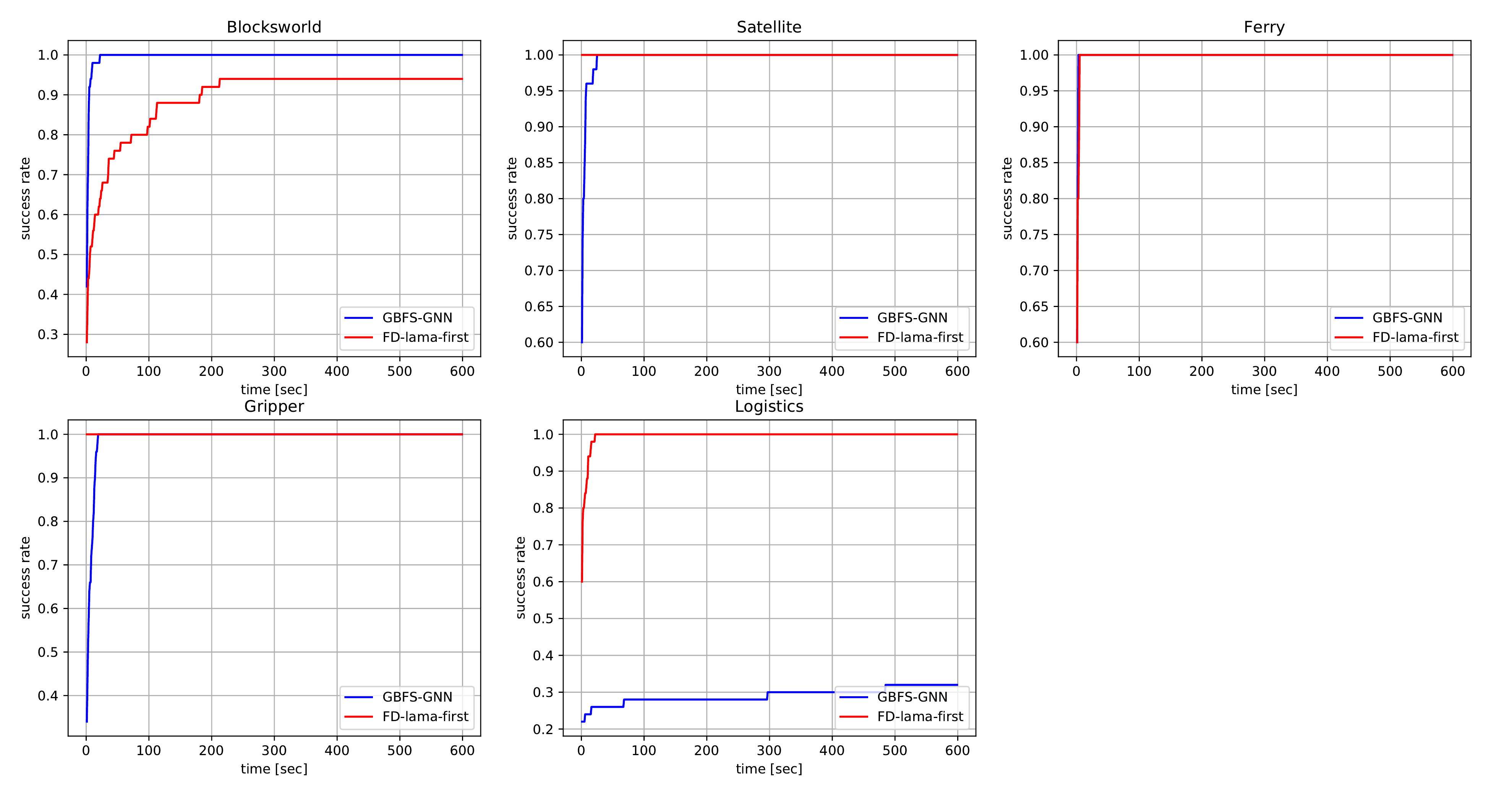}
        \caption{These plots compare success rate against run time for the various domains used in the evaluation}
        \label{fig:sr_time_results}
    \end{center}
\end{figure}

In figure \ref{fig:sr_time_results} we present a comparison between our method and fast-downward, plotting success rate against given run-time. We can see that even though fast-downward has a highly optimized C++ implementation and uses sophisticated modeling tools to efficiently solve planning problems, our method overcomes it in one domain (Blocksworld) and closely matches it on three others. Despite GBFS-GNN using a successor state and legal action generator that is orders of magnitude slower than that of fast-downward, our method's generalization capability makes it competitive with state of the art implementations of classical planners.

An obvious exception concerning the generalization performance of our method is the Logistics domain. Our policy successfully achieved good performance on the training instances but failed to generalize to much larger instance sizes, and consequently was vastly outperformed by fast-downward on that domain. We hypothesize that unlike the other domains, the Logistics domain contains a tighter coupling between the different objects in each instance. In the Satellite domain for example, calibrating an instrument or imaging a target does not interfere with other satellites, in the sense that the policy can have multiple "half-baked" goals and switch between them without interference. This is not possible in the Logistics domain, as all the packages share the trucks and airplanes, and moving a specific truck to pick up a package might interfere with another package that was meant to be picked up in another location. Different graph neural network architectures could perhaps  encourage the policy to remain "fixed" on a single goal until its satisfaction before moving to another, thus possibly overcoming the issue with the Logistics domain and other similar types of problems.

\section{Conclusion and Future Work}

In this work we studied the ability of graph neural networks and deep reinforcement learning algorithms to learn generalized planning policies that can solve instances much larger than those encountered during training, in effect learning principles that generalize well. Unlike some other approaches, our method does not rely on optimal solutions provided by existing planners, nor on heuristics to boost performance. We further introduce GBFS-GNN, a search algorithm that exploits the availability of high performing reactive policies to quickly find solutions to very large instances. Our policies are learned from scratch via reinforcement learning, and combined with GBFS-GNN achieve performance that surpasses highly optimized implementations of state of the art planners in terms of expanded states, and is on par in terms of run-time. 
Directions for future work include studying how specific mechanisms in graph neural networks architectures relate to the emergent generalization behaviour on different domains, studying the effect of different reinforcement learning algorithms on generalization and perhaps exploring regularization schemes on the policy training procedure that might encourage better generalization.

\bibliographystyle{plain}
\bibliography{references}
\end{document}